\documentclass[10pt,twocolumn,letterpaper]{article}

\usepackage[pagenumbers]{cvpr} 

\usepackage{graphicx}
\usepackage{amsmath}
\usepackage{amssymb}
\usepackage{booktabs}
\usepackage{adjustbox}
\usepackage{pifont}
\usepackage{multirow}

\usepackage[pagebackref,breaklinks,colorlinks]{hyperref}
\usepackage[dvipsnames]{xcolor}

\usepackage[capitalize]{cleveref}
\crefname{section}{Sec.}{Secs.}
\Crefname{section}{Section}{Sections}
\Crefname{table}{Table}{Tables}
\crefname{table}{Tab.}{Tabs.}

\newcommand{\cmark}{\ding{51}}
\newcommand{\xmark}{\ding{55}}

\newcommand{\modelname}{Retrieving-to-Answer}
\newcommand{\shortmodelname}{R2A}

\def\cvprPaperID{5117} 
\def\confName{ICCV}
\def\confYear{2023}

\begin{document}

\title{Retrieving-to-Answer: Zero-Shot Video Question Answering with \\ 
Frozen Large Language Models}

\author{Junting Pan$^{1,2}$\thanks{Equal contribution}\quad\quad Ziyi Lin$^{1*}$\quad\quad Yuying Ge$^{3}$\quad\quad Xiatian Zhu$^{4}$\quad\quad \\ 
Renrui Zhang$^{1,2}$\quad\quad Yi Wang$^{2}$\quad\quad Yu Qiao$^{2}$\quad\quad\quad Hongsheng Li$^{1}$\quad \vspace{.5em} \\
    $^1$Multimedia Lab, The Chinese University of Hong Kong, $^2$Shanghai AI Laboratory \\ $^3$University of Hong Kong, $^4$University of Surrey \\
}
\maketitle

\begin{abstract}

Video Question Answering (VideoQA) has been significantly
advanced from the scaling of recent Large Language Models (LLMs).
The key idea is to convert the visual information into the language feature space so that the capacity of LLMs can be fully exploited.
Existing VideoQA methods typically take two paradigms:
(1) learning cross-modal alignment,
and 
(2) using an off-the-shelf captioning model to describe the visual data.
However, the first design needs costly training on many extra multi-modal data,
whilst the second is further limited by limited domain generalization. 
%
%
To address these limitations, 
a simple yet effective {\bf Retrieving-to-Answer} (R2A) framework is proposed.
Given an input video, R2A first retrieves a set of semantically similar texts from a generic text corpus using a pretrained multi-modal model (\eg, CLIP). 
With both the question and the retrieved texts, 
a LLM (\eg, DeBERTa) can be directly used to yield a desired answer.
%
Without the need for cross-modal fine-tuning, R2A allows for all the key components (\eg, LLM, retrieval model, and text corpus) to plug-and-play. 
Extensive experiments on several VideoQA benchmarks show that despite with 1.3B parameters and no fine-tuning, our R2A can outperform the 61$\times$ larger Flamingo-80B model~\cite{flamingo} even additionally trained on nearly 2.1B multi-modal data.
\vspace{-3mm} 

\end{abstract}


\section{Introduction}
\label{sec:intro}

Video Question Answering (VideoQA) aims to answer a question regarding a reference video \cite{xu2017video}.
Due to the open-end nature,
manually annotating a large comprehensive dataset dedicated for VideoQA is 
practically impossible \cite{vlbert, ClipBERT, sun2019videobert}.
An appealing approach to address this challenge is {\em zero-shot learning} as pioneered by recent attempts \cite{flamingo,frozen,plugplay,frozenbilm,vidil}.
Instead of training a task-specific model,
they resort to learn a general-purpose multi-modal model using strong pretrained Large Language Models (LLMs) \cite{clip, blip, gpt3, deberta,instruct-gpt3,bert,flan}, because LLMs 
can accommodate rich knowledge from text data at scale.

\begin{figure}[!t]
\centering
\includegraphics[width=\linewidth]{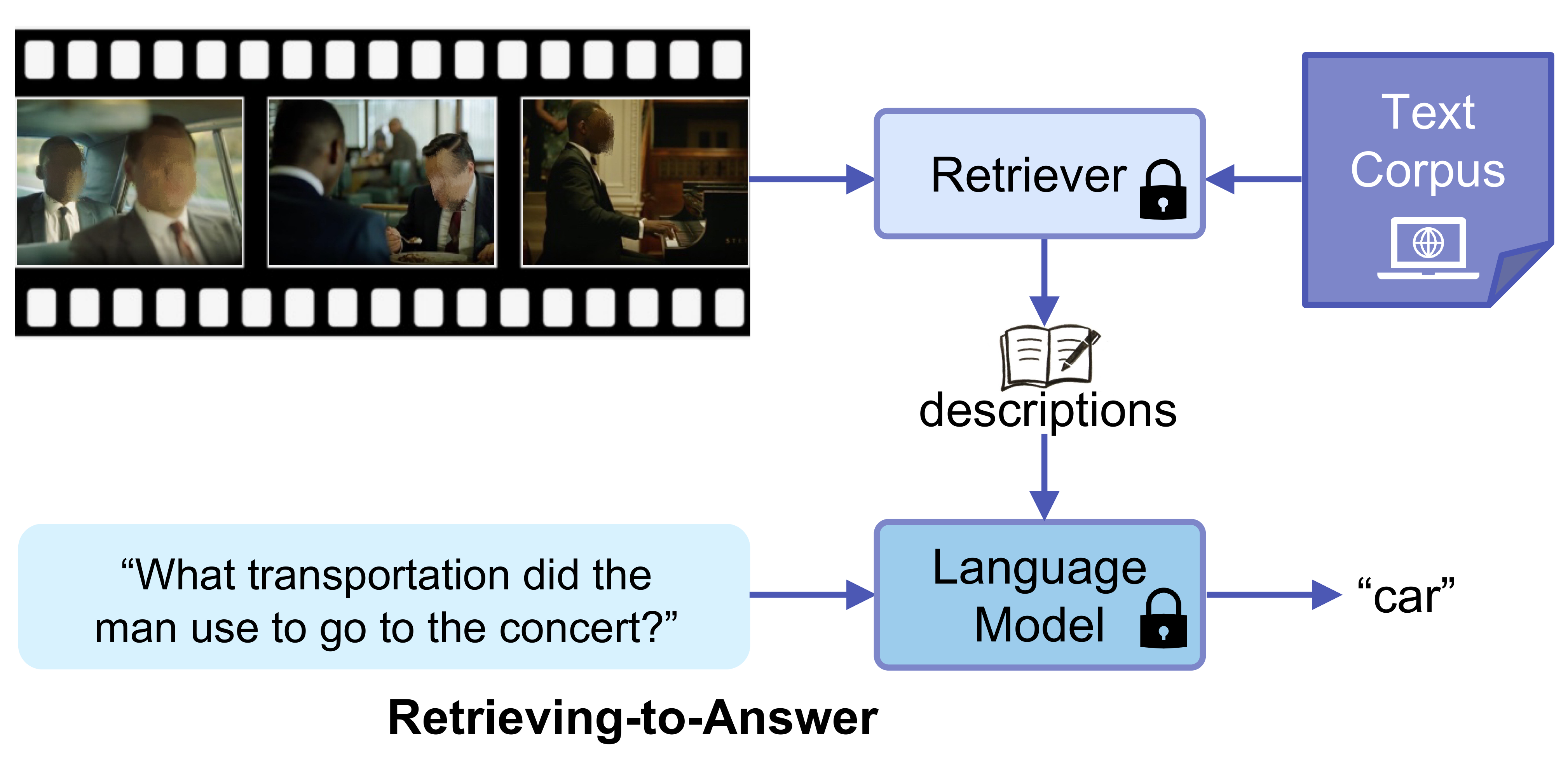}
\caption{
Overview of our {\bf Retrieving-to-Answer} (R2A) framework for zero-shot video question answering.
Given a reference video, R2A efficiently retrieves a set of semantically similar texts from an external corpus (\eg, WebVid~\cite{frozenintime}). 
With both the retrieved texts and the question,
a pretrained language model can be used directly to yield the final answer.
Without additional training, 
R2A allows any component to plug-and-play. 
}
\label{fig:figure1}
\vspace{-3mm}
\end{figure}

To capitalize off-the-shelf LLMs for VideoQA, the key lies in how to bridge the gap between texts and videos effectively.
There are two existing paradigms.
The {\bf\em first} adopts a cross-modal alignment strategy that projects visual features into soft prompts in the text embedding space. 
This design has two limitations:
(1) 
High training cost due to both large-sized model and data \cite{cc, frozenintime, merlot, yfcc, howto100m}.
(2) 
Less flexibility in component upgrading,
as changing any component requires model retraining.
To avoid both issues, the {\bf\em second} paradigm 
instead connects the vision and text modalities by using an off-the-shelf caption model to
convert the reference video into textual description \cite{vidil}.
However, 
this method relies on caption models finetuned 
towards the target domain 
making them less generalizable.

More broadly in image generation, instead of generating visual elements from scratch, leveraging relevant elements retrieved from a large image collection could facilitate the synthesis of complex scenes \cite{isola2013scene}.
Connecting with the captioning-based paradigm above,
we draw an analogy that rather than generating captions using a caption model, retrieving from a text corpus could be an alternative way to obtain related text descriptions for the video.
To that end,
a strong retriever and a comprehensive text corpus are needed.
For the former, we see the potential in recent tremendous advances in contrastive multi-modal models 
(\eg, CLIP) showing remarkable abilities in zero-shot cross-modal retrieval \cite{clip}.
The latter is generally available, \eg, large diverse texts from the internet.

Under the above analysis and insight, we propose a simple yet effective {\bf\em Retrieving-to-Answer} (R2A) framework.
Instead of costly video captioning, we resort to more efficient cross-modal text retrieval from a generic text corpus (\eg, uncurated web data WebVid~\cite{frozenintime}), simply leveraging a pretrained contrastive multi-modal model (\eg, CLIP~\cite{clip}).
Given an input video, we first retrieve a pool of semantically similar texts from the corpus in the multi-modal model's feature space.
This step can be considered as semantic video summarization. 
With the retrieved texts and the question, a pretrained large language model (\eg, DeBERTa \cite{deberta}) can be then directly applied to generate the answer.
R2A achieves state-of-the-art performance on multiple VideoQA benchmarks and simultaneously addresses the limitations of previous paradigms: (1) Our modular design allows our R2A to accommodate readily available pretrained models for the VideoQA task without the need for fine-tuning.
(2) Our R2A is able to generalize to new tasks/domains without further adaption because of both the selected powerful LLMs and multi-modal foundation model~\cite{bommasani2021opportunities}
(\eg, CLIP ~\cite{clip}
with remarkable generalization ability validated on novel domains and tasks~\cite{hu2023open, coop}),
and the usage of a large diverse text corpus.
Moreover, as the text corpus just needs to be encoded once,  the whole retrieval process of R2A is fast  (\eg, with naive implementation, retrieving 10 captions per video only takes 3.5ms on a 10M-sized text corpus). 

Our {\bf contributions} are summarized as follows: (I) We propose a novel idea of text retrieval for zero-shot video question answering, in contrast to previous cross-modal alignment learning and video captioning strategies.
(II) We introduce a simple, more efficient, yet more performing  {\em Retrieving-to-Answer} (R2A) framework, without additional fine-tuning whilst being fully open to the selection and change of any components.
(III) 
Extensive experiments show that our R2A achieves new state-of-the-art performance on multiple benchmarks in the zero-shot setting. In particular, with only 1.3B parameters and no additional tuning, our model outperforms the cross-modal training-based Flamingo~\cite{flamingo} with 80B parameters.

\section{Related Work}
\label{sec:relatework}

\subsection{Recent advances of VideoQA} 
Video Question Answering (VideoQA) has gained increasing attention due to its wide applications in video search, summarization, and understanding. 
%
%
Involving natural language comprehension, question answering, and video processing, this task presents a number of typical multi-modal learning challenges simultaneously.
Many prior methods rely on supervised model learning 
from labeled VideoQA datasets~\cite{choi2021dramaqa,dang2021object,kim2021self,lin2021vx2text,park2021bridge,sadhu2021video,seo2021look,xiao2021next,seo2022end,le2020hierarchical,xiao2021next,li2022invariant, zhong2022video,xiao2022video}.
Due to the limited size of manually labeled data,
the resulting models are less capable and generalizable across domains.
%
%
%
To mitigate this obstacle, 
recent methods adopt a paradigm of
first pretraining on large vision-language data and then fine-tuning on the target small training set~\cite{violet, allinone, ClipBERT, justask, merlot_reserve}. 
This approach still focuses on task-specific settings
with limited domains involved.
%
%
For more domain-generalizable VideoQA,
zero-shot learning has recently shown potential,
with the promising ability to scale to previously unseen samples with zero supervision \cite{justask, merlot, merlot_reserve, flamingo, frozenbilm}. 
%
For example, Reserve~\cite{merlot_reserve} learns to understand vision-language knowledge from web videos and the corresponding transcripts. Flamingo~\cite{flamingo} and FrozenBiLM~\cite{frozenbilm} are established using frozen pretrained models through cross-modal training. In contrast, our approach can leverage readily available pretrained models without the need for costly cross-modal training.

\begin{figure*}[t]
\centering
\includegraphics[width=\textwidth]{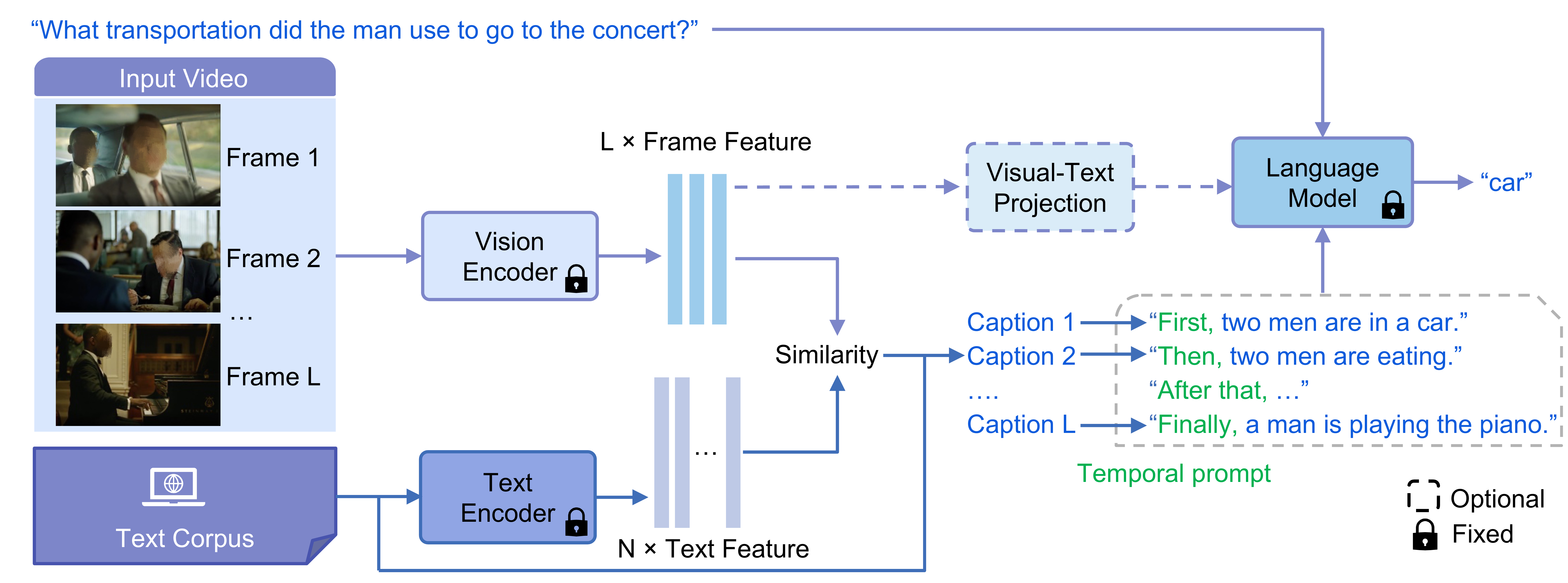}
\caption{ \textbf{Overview of our {\em \modelname{}} (R2A)~framework.} 
With the text encoder of a pretrained vision-language (ViL) model (\eg, CLIP \cite{clip}),
we first encode all the captions of an external text corpus (\eg, WebVid~\cite{frozenintime}) into the aligned representation space {\em for one time}.
Given a test video, 
{\bf(a)} we first extract the frame features using the vision encoder of the CLIP model. 
{\bf(b)} 
Subsequently, for each video frame, we retrieve top-$k$ semantically similar texts from the text corpus against the corresponding frame feature vector.
{\bf(c)} We then, combined the top-$k$ retrieved text for each frame with temporal prompts to construct a video-level textual context.
{\bf(d)}  
With both the question and the video-level context as input, a pretrained language model (\eg, DeBERTa \cite{deberta}) can be directly used to yield the final answer. 
{\bf(e)} {\em Optionally}, a visual-text projection layer can be learned to prompt the same language model, taking the video feature as input.
\label{fig:framework}
}
\vspace{-2mm}
\end{figure*}

\subsection{ LLMs for Vision and Language Tasks} Large Language Models (LLMs) have demonstrated impressive generalization capability on Natural Language Processing (NLP) tasks, thanks to their rich knowledge learned from vast text data \cite{gpt3,bert,deberta,instruct-gpt3,flan}. Recently, LLMs have been applied to vision and language (ViL) tasks, such as image captioning~\cite{visualgpt} and visual question answering~\cite{flamingo,frozenbilm}. 
Applying LLMs for ViL tasks is challenging due to the cross-modal gap. 
Importantly, LLMs are expensive to run,
let alone fine-tuning them on large target datasets.
For instance, the GPT-3~\cite{gpt3} model with 175B parameters requires 350GB of GPU memory to perform inference, not to mention training. 
Such high resource demands become practical obstacles for model training.
This motivates the development of cheaper methods to bridge vision and language. 
{\flushleft \bf Training-based adaption} Tsimpoukelli~\etal~\cite{frozen} train a vision encoder that encodes visual information into text embedding for cross-modality alignment while freezing the LLMs. Flamingo~\cite{flamingo} incorporates new cross-attention layers into existing frozen LLMs during training. FrozenBiLM~\cite{frozenbilm} achieves state-of-the-art performance on zero-shot VideoQA by adapting frozen bidirectional Language Models~\cite{bert}. 
MAPL~\cite{mapl} and VisualGPT~\cite{visualgpt} also leverage  cross-modal fine-tuning of large pre-trained models, with focus on VQA and image caption tasks.
Commonly, these methods all require joint training of vision and language models together, which is computationally expensive due to large scales of both the model~\cite{gpt3, flan} and dataset size~\cite{cc, frozenintime, merlot, yfcc, howto100m}. For example, Flamingo~\cite{flamingo} was trained for 500K TPU hours on billions of image/video-text pairs. Despite that approaches like FrozenBiLM~\cite{frozenbilm} explored ways to mitigate this issue, the problem is still far from being solved (requiring up to hundreds of GPU hours and millions of video-text pairs for adaptation). 
{\flushleft \bf Language-based adaption} Instead of training modality alignment modules, some recent approaches leverage language directly for connecting visual cues with LLMs~\cite{plugplay, socratic, pica, vidil}. Specifically, they all use an image-caption model to generate textual descriptions of visual information. While there is no need for training-based cross-modal alignment, they additionally suffer from the limitations of the caption module used, such as limited domain knowledge and high inference cost. (which is in the same domain as the target VQA tasks) for caption generation, PnP-VQA~\cite{plugplay} must go through multiple processing steps to generate captions for achieving good performance on visual question answering, resulting in high inference cost. 
In contrast, our approach retrieves text descriptions from external world knowledge using a generic multi-modal retriever~\cite{clip},
much more efficient than captioning whist no extra limitations. 
Note, our retrieval process can be implemented easily with highly optimized open-source libraries (\eg, SCaNN~\cite{scann} can query 2 trillion tokens in $10ms$).

\subsection{Retrieval augmentation}
Retrieval-augmented language models have gained attention in NLP~\cite{retro, realm, wu2022memorizing, lewis2020pre, lewis2020retrieval, lewis2020retrieval}. In vision-related applications, there are attempts at exploiting semantically related examples for improving model training or inference, \eg\ visual question answering \cite{izacard2020leveraging,lewis2020retrieval}, image captioning \cite{zhang2021open,sarto2022retrieval,ramos2022smallcap}, recognition~\cite{iscen2022memory,long2022retrieval}, synthesis~\cite{qi2018semi,isola2013scene}, human parsing~\cite{liu2015matching}, and many more ~\cite{retro,realm,wu2022memorizing,lewis2020pre}. Nagrani \etal~\cite{nagrani2022learning} transfer image captions to video according to visual similarity for building video-language datasets. 
VideoCLIP~\cite{videoclip} adopts retrieval-augmented training for mining hard negative samples.
Our work belongs to this line of research.
To the best of our knowledge, this is the first attempt that exploits the strategy of cross-modal retrieval from open-world knowledge for the VideoQA tasks.

\section{Methodology}

For zero-shot VideoQA, 
we formulate a {\bf \modelname} (\shortmodelname) framework
by efficiently bridging the readily available pretrained vision and language models  (\eg, CLIP~\cite{clip} and DeBERTa~\cite{deberta}).
The overall architecture is depicted in Figure~\ref{fig:framework}. Via retrieval-based cross-modal search in an external text corpus (\eg, WebVid \cite{frozenintime}), we transfer the information of video into text; As a result, an existing language model can be applied to VideoQA.

We first describe how to encode video (Section~\ref{method:vid_enc}) and text (Section~\ref{method:txt_enc}). Then, we explain in Section~\ref{method:vidtxt_ret} how to retrieve semantically relevant textual context given a query video. We present in Section~\ref{method:answer} how we answer the question with the retrieved context. Finally, we describe in Section~\ref{method:ft} how to (optionally) further learn a visual-to-text layer for a further performance boost.

\subsection{Text Corpus Encoding}
\label{method:txt_enc}

Our \shortmodelname{} is charaterized with an external generic text corpus 
for contextual text retrieval. 
We use the text encoder $\texttt{E}_{\texttt{Text}}$ of the same ViL model (CLIP in this case)
to text vectorization.
Suppose there are $N$ samples in the corpus, with each consisting of a word sequence $T_i$. 
For efficient retrieval, we precompute the textual features for all text samples: $v_i = \texttt{E}_{\texttt{Text}}(T_i) \in \mathbb{R}^{d}$, with $i=1,...,N$.
This process takes place once.
In the case of CLIP, similarly $v$ corresponds to the feature of the \texttt{[CLS]} token.

\subsection{Video Encoding}
\label{method:vid_enc}

For video encoding, we use the vision encoder $\texttt{E}_{\texttt{Vis}}$ of a pretrained ViL model (\eg, CLIP \cite{clip}).
Suppose we have an input video $V =[f_1, f_2, ..., f_L]$, where $L$ is the number of frames.
We extract a visual feature embedding for each frame of the video using the encoder,
denoted as: $z_t = \texttt{E}_{\texttt{Vis}}(f_t)\in \mathbb{R}^{d}$, with $t=1,...,L$.

In practice, we adopt CLIP’s ViT-L/14 variant whilst other similar models such as ALIGN \cite{jia2021scaling} can be similarly considerable.
The frame feature embedding
$z$ corresponds to the output feature of the \texttt{[CLS]} token.

\subsection{ Video-Text Retrieval} 
\label{method:vidtxt_ret}

We obtain the contextual text by video-text retrieval from the external text corpus. 
Specifically, we compute the similarity score at video frame level between frame feature $z_{t}$ and text feature $v_i$:
\begin{equation}
d(z_{t}, v_i) ={z_{t} \cdot v_i \over \| z_{t} \|\|v_i \|}
\end{equation}
where $\cdot$ defines the dot product operation.
This is the default choice in our main experiments.

Given all the pairwise similarity scores, we rank 
the corpus samples in the descending order. 
The contextual text for each frame is obtained as the top $k$ matches,
denoted as $r_t = [T_{t_1},..., T_{t_k}]$, where $t_k$ stands the for top $k$-th text match for the $t$-th frame. The contextual text for the whole video is denoted as: $R = [r_1,..., r_L]$.

\subsection{Temporal Aware Prompting} 
\label{method:answer}
In order to capture temporal transitions between frames, we further process the retrieved video contextual text. Specifically, similar to~\cite{vidil} we add temporal aware prompts in natural language indicating the temporal order of the retrieved captions, as a result, the prompted contextual text has the following form: \textcolor{teal}{``\texttt{Firstly,\{$r_1$\}... Then,\{$r_t$\}... After that,... Finally,\{$r_L$\}}"}.

\subsection{Answer Generation} 
\label{method:answer}
To generate an answer,
we exploit a pretrained language model
conditioned on both the question and the contextual text we retrieve as above.
As a showcase, in practice we use a BERT~\cite{bert} style language model pretrained with the Masked-Language-Modeling (MLM) task. 
Concretely, the objective of MLM is to predict the values of all masked tokens
given a word sequence under random masking.
For example, given an input sequence as ``\texttt{Paris [MASK] the [MASK] city of France}'', 
via contextualizing over visible word tokens, the model can predict 
the two masked tokens ``\texttt{is}'' and ``\texttt{capital}''.
To fit this scheme, we construct a prompt template as: 
\textcolor{magenta}{``\texttt{Question:\{question\} Answer:[MASK] Hints:\{prompted contextual text\}}"}.
We use this template to prompt the pretrained language model
which could 
output the answer at the designated \texttt{[MASK]} position. For the example in Fig.~\ref{fig:framework}, the input text for the LM will be like this: ``\texttt{Question: What transportation did the man use to go to the concert? Answer: [MASK]. Hints: First, two men are in a car.... Finally, a man is playing piano.}''

\subsection{\shortmodelname{} with Learnable Visual-Text Projection } 
\label{method:ft}
We have discussed our standard R2A framework as above.
In cases that we have access to a video-text pair training data (\eg, WebVid~\cite{frozenintime}, not VideoQA specific training data since we consider the zero-shot setting here)
and aim to pursue further performance boost, a lightweight learnable module can be simply integrated on top (see Figure~\ref{fig:framework}). We denote the fine-tuned version of R2A as \textbf{R2A-FT}.
However, we note that this will discard a certain degree of flexibility 
as this extra training is coupled with other pre-trained vision and language models.

For training cost minimization, we only learn a {\em visual-to-text projection layer} to map the video features to prompts,
whilst freezing the language model. 
The video-conditioned prompts are learned to be compatible with the language model.
Formally, this component can be written as:
\begin{equation}
    p_{vid} = \{z_t\mathbf{W}_{proj}\}_{t=1}^L
\end{equation}
where $\mathbf{W}_{proj}$ denotes the learnable parameters with our visual-to-text projection layer. 
It is randomly initialized, and the only part to be optimized.
Different from existing alternatives \cite{frozenbilm}, we do not include any adapters in the frozen language model.

Concretely, the model takes as input the retrieved contextual text (captions), the video-conditioned prompts, and the original caption with the video. Model training is conducted by the MLM task:
\begin{small}
\begin{equation}
    \texttt{log}\ p(y|p_{vid}, R, T_{vid}) = \sum_{m=1}^{M} \texttt{log}\ p(y_m|p_{vid},R, T_{vid})
\end{equation}
\end{small}
where $T_{vid}$ is the original caption from the training set with tokens randomly masked out, $y$ are the values of those masked tokens and $M$ is the number of masked tokens.

We set the mask ratio to 50\%. 
Following the original implementation of BERT~\cite{bert}, we replace the masked token position with the \texttt{[MASK]} token at the probability of 80\%, with a random token at 10\%, and with the original token at 10\%. During inference, we follow the same manner as discussed in Section~\ref{method:answer}, except that we further prepend the newly generated video-conditioned prompts.  

\section{Experiments}

\begin{table*}[t]
    \centering
    \adjustbox{width=1.0 \linewidth}{
    \begin{tabular}{@{}l|cc|cc|ccccccc@{}}
        \toprule
        \multirow{2}{*}{Method} & \multicolumn{2}{c|}{Language} & \multicolumn{2}{c|}{Vision} & \multicolumn{6}{c}{Benchmarks}  \\
         & model & \#params & model & \#params & MSRVTT-QA & MSVD-QA & ANet-QA & TGIF-QA & iVQA & NextQA\\ 
        \midrule
        \multicolumn{5}{@{}l}{\it Training based Adaption} \\
        CLIP ViT-L/14 \cite{clip}     & Custom & 123M & ViT-L/14 & 300M & 2.1 & 7.2 & 1.2 & 3.6 & 9.2 & -\\
        Just Ask \cite{justask}       & DistilBERT & 66M & S3D & 12M & 5.6 & 13.5 & 12.3 & - & 13.3 & -\\
        Reserve \cite{merlot_reserve} & Custom & - & ViT-L/16 & 300M & 5.8 & - & - & - & -  & -\\
        Flamingo-3B \cite{flamingo}   & Chinchilla-like & 2.6B  & NFNet-F6 & 629M & 11.0 & 27.5 & - & - & 32.7 & 21.3\\
        Flamingo-9B \cite{flamingo}   & Chinchilla-like & 8.7B  & NFNet-F6 & 629M & 13.7 & 30.2 & - & - & 35.2 & 23.0\\
        Flamingo-80B \cite{flamingo}  & Chinchilla-like & 80B   & NFNet-F6 & 629M & 17.4 & 35.6 & - & - & \textbf{40.7} & 26.7\\
        FrozenBiLM \cite{frozenbilm}  & DeBERTa-v2-XL   & 890M  & ViT-L/14 & 300M & 16.9 & 33.7 & 25.9 & 41.9 & 26.2 & - \\
        \midrule
        \multicolumn{5}{@{}l}{\it Language based Adaption} \\
         VidIL* \cite{vidil} & DeBERTa-v2-XL & 890M  &ViT-L/14 & 300M & 16.6 & 31.7 & - & - & - & - \\
        R2A (Ours)    & DeBERTa-v2-XL   & 890M  & ViT-L/14 & 300M & \textbf{18.3} & {\bf 37.0} & {\bf 26.3} & \textbf{52.2} & 29.3 & \textbf{34.7}  \\
        \bottomrule
    \end{tabular}
    } 
    \vspace{-2mm}
    \caption{{\bf Comparison with state-of-the-arts on zero-shot videoQA.} 50 retrieved sentences are used per video and the prompt is ``{\tt Hints:}''. *indicates replacement of LM beyond original definitions by authors for fair comparisons and also due to a lack of access to the original Language Model (GPT-3).
    } 
    \label{tab:sota}
\end{table*}

We first describe the VideoQA datasets used for our evaluations (Section~\ref{subsec:data}). \
We then give the implementation details of our model (Section~\ref{subsec:implementation}). Next, we compare our R2A with the state of the art methods
(Section~\ref{subsec:sota}). 
Finally, we ablate the effect of different components, \eg, the choice of pretrained models, the number of retrieval samples and the construction of the text corpus (Section~\ref{subsec:ablation}).

\subsection{Datasets and Evaluation} 
\label{subsec:data}
In this work, we focus on the challenging open-ended VideoQA datasets where the model has to generate an open-ended answer for each video-question pair. We test multiple zero-shot VideoQA benchmarks with data collected from sources of great diversity  (\ie, YouTube videos, Sports Videos, GIFs), including \textbf{MSRVTT-QA}~\cite{msrvtt}, \textbf{MSVD-QA}~\cite{msrvtt}, \textbf{ActivityNet-QA}~\cite{activitynet-qa}, \textbf{TGIF-QA}~\cite{tgif}, \textbf{iVQA}~\cite{justask} and \textbf{NextQA}~\cite{xiao2021next}.
We report top-1 accuracy based on exact matching between the predicted answer and ground-truth annotation following previous evaluation protocols~\cite{allinone}. 
\begin{figure*}
    \centering
    \includegraphics[page=1,width=0.32\textwidth]{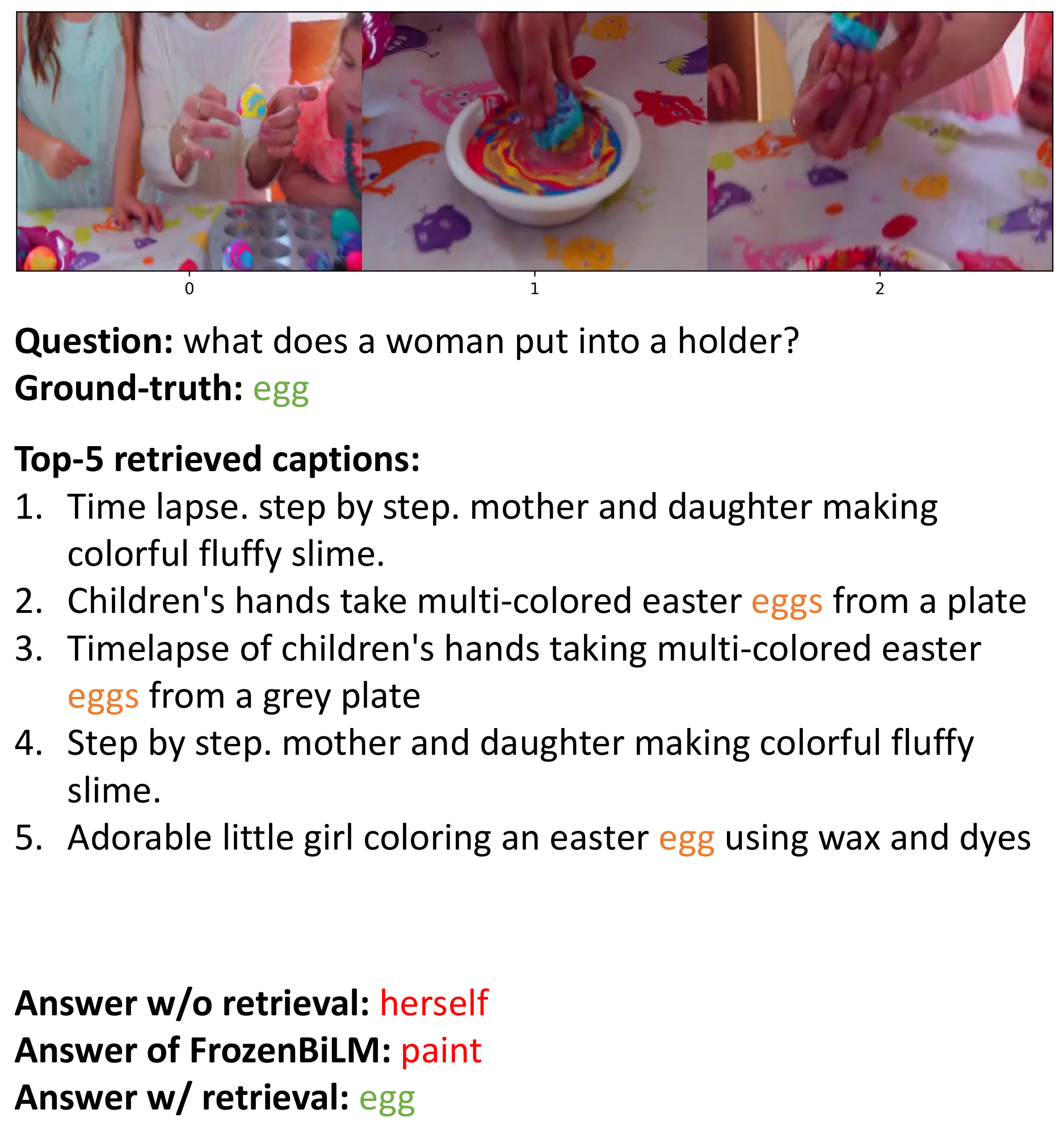}
    \includegraphics[page=2,width=0.32\textwidth]{figs/msrvtt_vis.pdf}
    \includegraphics[page=3,width=0.32\textwidth]{figs/msrvtt_vis.pdf}
    \vspace{-3mm}
    \caption{{\bf Qualitative comparison of model predictions with and without retrieved captions.}  For illustrative purposes, we highlight words in \textcolor{orange}{orange} to indicate  answer cues from captions. Words in \textcolor{YellowGreen}{green} indicate correct answer predictions and in \textcolor{red}{red} for incorrect ones.}
    \vspace{-3mm}
    \label{fig:vis}
\end{figure*}

For the external corpus, we consider text extracted from various multi-modal datasets, including two datasets scraped from web: (1) \textbf{WebVid}~\cite{frozenintime} that contains approximately 10M video-text pairs,
(2) \textbf{CC3M}~\cite{cc} and \textbf{CC12M}~\cite{cc12m} that consists of 3M and 12M image-text pairs, as well as a human-annotated dataset: (3) \textbf{COCO Caption} dataset~\cite{coco} with 1.5M human-generated captions describing over 330K images. 

%

\subsection{Implementation details}
\label{subsec:implementation}

For video-to-text retrieval (sec.~\ref{method:vidtxt_ret}), we use the ViT-L/14 variant of CLIP~\cite{clip}. For the LLMs, we adopt DeBERTa-XL~\cite{deberta} as our default language model. 
It is worth mentioning that we utilize the MLM pre-trained model checkpoint, which means it has never been trained on QA-related tasks in any modality. Unless stated otherwise, we set 500 as the maximum input length for our language models. We base our implementation on the officially released code of \cite{frozenbilm} and the unmentioned details follow their implementation. 

\vspace{-3mm}

\paragraph{Visual feature extraction} We extract frame features using the ViT-L/14 variant of CLIP~\cite{clip}. The frame preprocessing follows the official implementation of \cite{clip}. We uniformly sample 10 frames from each video and extract one feature vector for each frame by taking the output of the $\texttt{[CLS]}$ token. 

\vspace{-3mm}

\paragraph{Video-to-Text Retrieval} The feature similarity calculation is identical to the original CLIP implementation. We use the naive algorithm for the nearest neighbor search (\ie, calculating the similarity between all pairs and selecting the top-$k$ for each query).  
%
More advanced nearest neighbor search algorithms (\eg \cite{guo2020accelerating}) can be used for larger datasets.  Duplicate entries are removed from the retrieved set.

\vspace{-3.5mm}

\paragraph{Multi-modal Fine-Tuning} 
For the multi-modal fine-tuning paradigm, we learn a linear projection for visual features as stated in Section \ref{method:ft}. We use an Adam optimizer with a constant learning rate of 1.5e-5, {\it no} weight decay and $\beta_1, \beta_2 = 0.9, 0.95$. By default, we use a batch size of 64, limit the input sequence length to 64 for training, and train for only one epoch on each dataset. We alter each input token with a probability of 0.5.

\subsection{Comparison with the State-of-the-arts}
\label{subsec:sota}

\begin{figure*}[t!]
    \centering
    \includegraphics[width=0.24\textwidth]{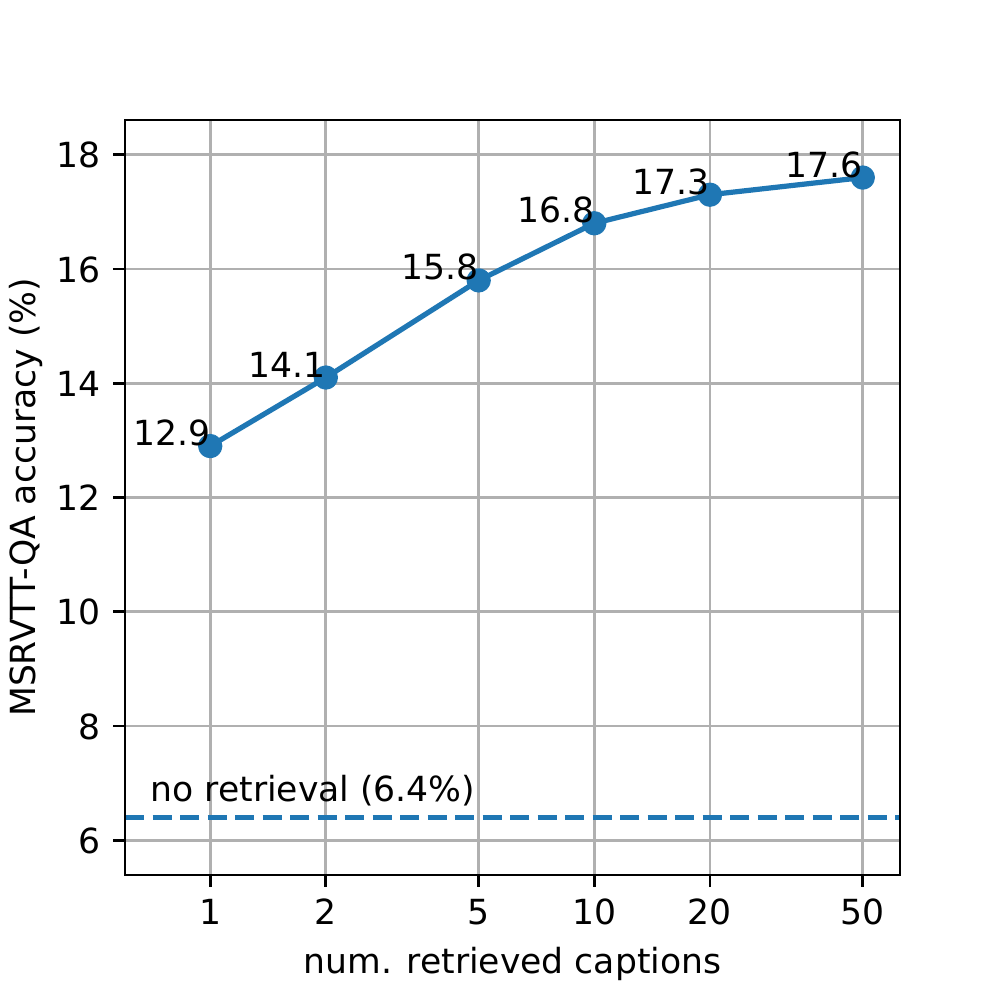}
    \includegraphics[width=0.24\textwidth]{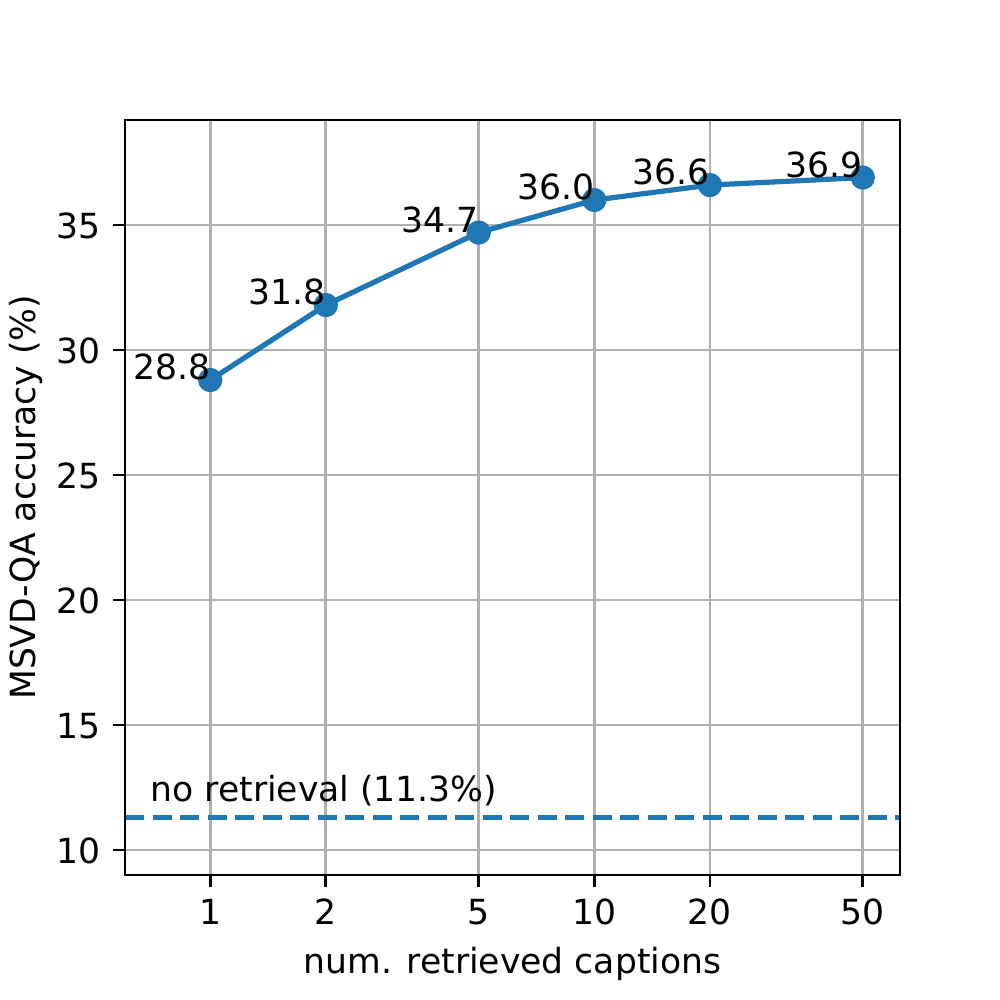}
    \includegraphics[width=0.24\textwidth]{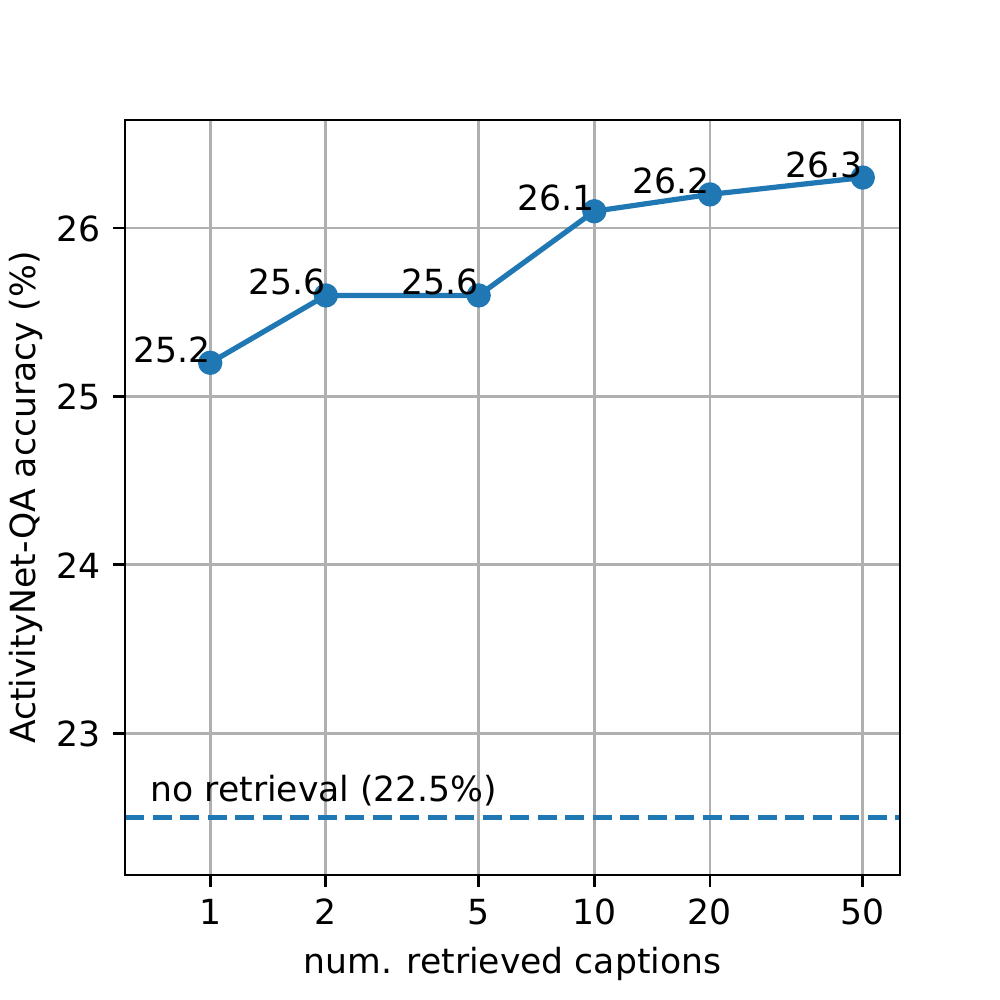}
    \includegraphics[width=0.24\textwidth]{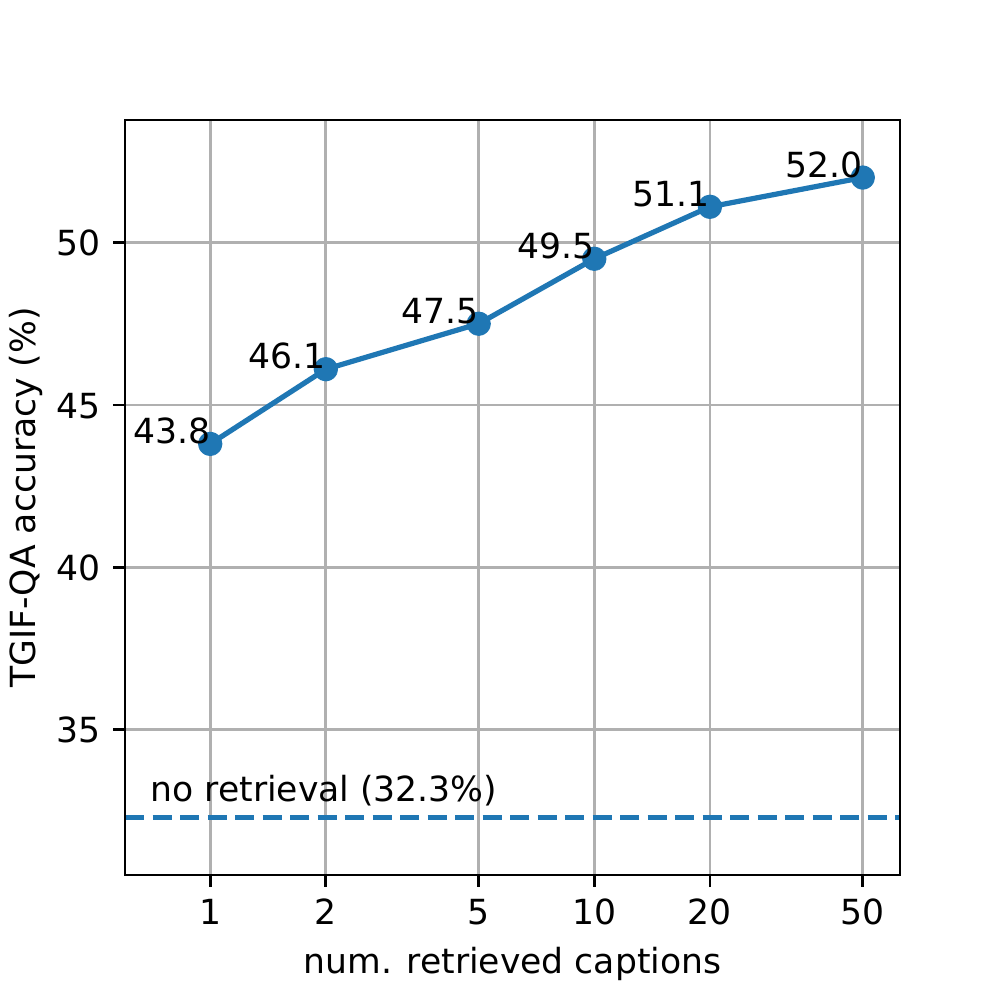}
    \vspace{-2mm}
    \caption{{\bf Effects of the number of retrieved captions per video on zero-shot VideoQA.} WebVid-10M is used as the retrieval set and ``{\tt Subtitles:}'' is used as the prompt in all experiments. }
    \vspace{-3mm}
    \label{fig:num_retrieve}
\end{figure*}

\paragraph{Quantitative comparison} Table~\ref{tab:sota} presents the results of our method in comparison with current state-of-the-art approaches on zero-shot VideoQA. Our method outperforms approaches that were additionally trained on million to billion-scale vision-language data, except on iVQA, where our method underperforms Flamingo~\cite{flamingo}. However, it is worth noting that Flamingo, while trained with billion-scale data, even the smallest version has significantly more parameters than our method.
Furthermore, when using the same language model, our method consistently improves over VidIL~\cite{vidil}, which utilizes a caption model to connect video with the language model. These results confirm the effectiveness of our method, as well as the informativeness of the retrieved captions for VideoQA.

\vspace{-3.5mm}

\paragraph{Qualitative comparison} Figure~\ref{fig:vis} illustrates qualitative results of zero-shot VideoQA for our \modelname~in comparison to FrozenBiLM~\cite{frozenbilm} and the text-only baseline without access to visual information (w/o retrieval). First, we observe that for questions only baseline, the LM can predict answers based on commonsense reasoning, \eg in the second example, it is very likely for a teacher to write problems on paper if we do not consider the visual input.  Second, for FrozenBiLM, predictions can be misled by inaccurate visual information (\eg in the first example, it predicts ``paint'' instead of ``egg''). In contrast, our R2M can predict the correct answer based on high-quality informative context retrieved from the supportive text corpus.

\vspace{-3.5mm}

\paragraph{Efficiency analysis} As shown in Table~\ref{tab:efficiency}, the inference time per video for our R2A is 0.11s, which can be further broken down into video encoding latency, retrieval latency, and LM inference latency. For FrozenBiLM, the inference time is 0.07s, which is composed of video encoding latency and LM inference latency. Given the exact same video input setup, the video encoding time is identical for our method and FrozenBiLM. Our LM inference latency is slightly larger than FrozenBiLM's because our input sequence (including the retrieved contextual text) is longer. 
It is worth noting that, the time consumed by retrieval is only 3.5ms (2,800 queries per second on a single A100 GPU from 10M samples) which is almost negligible compared to the total inference time. In addition to the inference time, we also report the pre-inference preparation time which is cross-modal training time for Flamingo and FrozenBiLM. In the case of R2A, it is the time needed to extract the features of the retrieval set which is fractional compared to the time required for cross-modal training. Overall, our results demonstrate that R2A achieves competitive performance while maintaining efficient inference time and requiring minimal pre-inference preparation.

\begin{table}[t]
    \centering
    \adjustbox{width=0.9\linewidth}{
    \begin{tabular}{@{}lccc@{}}
        \toprule
        Method & \begin{tabular}{@{}c@{}}Pre-inference\\computation\end{tabular} & \begin{tabular}{@{}c@{}} Inference\\ time per video\end{tabular} \\
        \midrule
        Flamingo-80B \cite{flamingo} & 500Kh (TPUv4) & - \\
        FrozenBiLM \cite{frozenbilm} & 160h (V100) & 0.07s \\
        \midrule
        R2A (Ours) & 1.4h (V100) & 0.11s \\
        \bottomrule
    \end{tabular}
    }
    \vspace{-1mm}
    \caption{{\bf Efficiency comparison.} We compare efficiency with two previous methods. {\it Pre-inference} includes the cross-modal training cost for training based methods and retrieval set feature extraction cost for ours. We report efficiency using 10 retrieved sentences per video, at which {\it R2A} outperforms FrozenBiLM on all tasks.}
    \label{tab:efficiency}
\end{table}


\subsection{Ablation Studies}
\label{subsec:ablation}

\begin{table}[]
    \centering
    \adjustbox{width=\linewidth}{
    \begin{tabular}{@{}lcccc@{}}
        \toprule
        Ret. modality & MSRVTT-QA & MSVD-QA & ANet-QA & TGIF-QA \\
        \midrule
        Video-Video & 15.7 & 32.6 & 25.0 & 45.2 \\
        Video-Text & {\bf 18.3} & {\bf 37.0} & \bf{26.3} & \bf{52.2} \\
        \bottomrule
    \end{tabular}
    } 
    \vspace{-1mm}
    \caption{{\bf Video-Text vs. Video-Video retrieval.} 50 retrieved sentences are used per video and the prompt is ``{\tt Hints:}''.}
    \label{tab:v2v_vs_v2t}
\end{table}

\begin{table}[]
    \centering
    \adjustbox{width=\linewidth}{
    \begin{tabular}{@{}ccccc@{}}
        \toprule
        ret.? & MSRVTT-QA & MSVD-QA & ANet-QA & TGIF-QA \\
        \midrule
        \multicolumn{5}{@{}l}{\it Language model: BERT-Base} \\
        \xmark & 1.4 & 2.9 & 16.0 & 10.4 \\
        \cmark & 3.0 (+1.6) & 7.5 (+4.6) & 15.4 (-0.6) & 17.4 (+7.0) \\
        \midrule
        \multicolumn{5}{@{}l}{\it Language model: BERT-Large} \\
        \xmark & 2.5 & 3.9 & 15.6 & 13.9\\
        \cmark & 7.0 (+4.5) & 15.0 (+11.1) & 19.6 (+4.0) & 26.7 (+12.8) \\
        \midrule
         \multicolumn{5}{@{}l}{\it Language model: RoBERTa-Large} \\
        \xmark & 3.0 & 5.5 & 18.5 & 13.2 \\
        \cmark & 13.6 (+10.6) & 24.3 (+18.8) & 18.8 (+0.3) & 32.0 (+18.8) \\
        \midrule
        \multicolumn{5}{@{}l}{\it Language model: DeBERTa-v2-xlarge (default)} \\
        \xmark & 6.4 & 11.3 & 22.5 & 32.3  \\
        \cmark & 16.8 (+10.4) & 36.0 (+24.7) & 26.1 (+3.6) & 49.5 (+17.2) \\
        \bottomrule
    \end{tabular}
    } 
    \vspace{-1mm}
    \caption{{\bf Performance of the training-free setting with alternative language models.} WebVid-10M is used as the retrieval set and 10 sentences are retrieved for each video. The prompt is ``{\tt Subtitles:}''. ``ret.?'' denotes whether to use retrieval or not.}
    \label{tab:lm}
    \vspace{-4mm}
\end{table}

\begin{table}[h]
    \centering
    \adjustbox{width=\linewidth}{
    \begin{tabular}{l@{}cccc@{}}
        \toprule
        Retrieval database & MSRVTT-QA & MSVD-QA & ANet-QA & TGIF-QA \\
        \midrule
        \multicolumn{5}{@{}l}{\it No retrieval dataset (baseline)} \\
        N/A & 6.4 & 11.3 & 22.5 & 32.3 \\
        \midrule
        \multicolumn{5}{@{}l}{\it Down-sampling the retrieval dataset} \\
        WV-10k & 15.6 & 30.3 & 25.0 & 44.2 \\
        WV-100k & 16.2 & 34.1 & 25.1 & 47.8 \\
        WV-1M & 16.3 & 35.6 & 25.8 & 48.5\\
        WV-10M & 16.8 & {\bf 36.0} & {\bf 26.1} & 49.5 \\
        \midrule
        \multicolumn{5}{@{}l}{\it Using the Conceptual Captions (CC) datasets} \\
        CC-3M & 14.2 & 30.2 & 25.0 & 47.1 \\
        WV-10M + CC-3M & 15.9 & 34.7 & 26.0 & 49.5 \\
        CC-12M & 10.9 & 23.6 & 23.4 & 43.2 \\
        WV-10M + CC-12M & 13.8 & 31.6 & 25.6 & 46.4 \\
        \midrule
        \multicolumn{5}{@{}l}{\it Using human-annotated captioning datasets} \\
        COCO & 16.4 & 31.2 & 23.9 & 45.3 \\
        COCO + WV-10M & {\bf 16.9} & {\bf 36.0} & 26.0 & {\bf 49.7}  \\
        \bottomrule
    \end{tabular}
    } 
    \vspace{-1mm}
    \caption{{\bf Effects of retrieval database construction.} 
    WV stands for the WebVid dataset \cite{frozenintime}. We use 10 retrieved sentences per video and the prompt ``{\tt Subtitles:}'' in all experiments. Note that we only use the textual part of each of the dataset. 
    }
    \label{tab:base_set}
\end{table}

\begin{table}[h]
    \centering
    \adjustbox{width=\linewidth}{
    \begin{tabular}{@{}l@{}cccc@{}}
        \toprule
        Retrieval database & MSRVTT-QA & MSVD-QA & ANet-QA & TGIF-QA \\
        \midrule
        N/A & 6.4 & 11.3 & 22.5 & 32.3 \\
        \midrule
        CC-12M (Random) & 7.7 & 10.0 & 19.6 & 30.1 \\
        CC-12M (Ours) & 10.9 & 23.6 & 23.4 & 43.2 \\
        \midrule
        WV-10M (Random) & 9.5 & 18.5 & 22.3 & 31.4 \\
        WV-10M (Ours) & 16.8 & 36.0 & 26.1 & 49.5 \\
        \bottomrule
    \end{tabular}
    } 
    \vspace{-2mm}
    \caption{{\bf Retrieval vs. random sample.} {\tt Subtitles:} is used as prompt and 10 sentences are used per video in all experiments.}
    \label{tab:retrieve_vs_random}
    \vspace{-4mm}
\end{table}

\paragraph{Retrieval dataset construction} Table~\ref{tab:base_set} presents the results of our method equipped with different text corpora for context retrieval. To investigate the impact of dataset size, we conduct experiments with various sample sizes. As shown in the second section of Table~\ref{tab:base_set}, the model's performance consistently improves with the increase in dataset size. Even a small dataset of 10k samples can bring a noticeable improvement over models without retrieved captions. We observe that the performance gain tends to be smaller when using CC-3M and CC-12M as the retrieval set, which we attribute to the dataset quality issues mentioned in \cite{mapl}. Specifically, we find some texts in the CC datasets unreadable, which can negatively affect the model's performance. Our experiments on COCO Captions reaffirm this hypothesis, as we observe higher performance gains for high-quality captions annotated by humans, even with much smaller dataset sizes. Unless otherwise stated, we use the full WebVid-10M as the default retrieval dataset.

\vspace{-3.5mm}

\paragraph{Impact of the number of retrieved captions}
We plot the top-1 accuracy against the number of retrieved captions, as shown in Figure~\ref{fig:num_retrieve}. The accuracy on all four datasets consistently increases as more retrieved captions are used, demonstrating the robustness of our method. Notably, we observe a substantial performance boost even when using only one retrieved caption, indicating the importance of the retrieval operation in assisting VideoQA. Furthermore, we find that the accuracy continues to improve with an increasing number of retrieved captions. These findings suggest that our method can effectively leverage external sources of information for VideoQA task.

\vspace{-3.5mm}

\paragraph{Impact of the quality of retrieved captions} To demonstrate the effectiveness of the retrieval in providing relevant information for VideoQA, we compare our results with those obtained by randomly sampling text to feed into the LM. As shown in Table~\ref{tab:retrieve_vs_random}, our R2A consistently outperform random sampling on all datasets and retrieval sources by a large margin. Surprisingly, we observe that randomly sampled text show improved performance over the question-only baseline on MSRVTT and MSVD. We hypothesize that this is due to the prior distributions of some retrieval datasets, which may serve as task-specific prompts \cite{coop} for certain target datasets.

\vspace{-3.5mm}

\paragraph{Video-Text vs. Video-Video Retrieval}
Apart from using only text as the external corpus, we experiment with retrieving video-text pairs: For a video-text dataset, we find the samples with the highest video-video similarity and take the corresponding text as the retrieved captions. As shown in Table~\ref{tab:v2v_vs_v2t}, the video-video retrieval is significantly worse on all four datasets.
We conjecture that video-video similarity is more vulnerable to data noise (\ie, some video-text pairs themselves may not be well aligned, especially for those web-crawled data), which consequently makes the retrieved text not well correlated with the query video.
Moreover, retrieving directly from the text corpus is also advantageous in term of storage (\ie, no need to store the images or videos) and flexibility (\ie, able to use both visual-text and text-only data).

\vspace{-3.5mm}

\paragraph{Impact of pretrained language models} We also experiment with alternative pretrained language models and report our results in Table~\ref{tab:lm}. Except for one case (BERT-base on ActivityNet-QA), we observe that the retrieved contexts significantly improve performance, sometimes even doubling the accuracy on all LMs. Notably, our models benefit more from stronger language models, as we observe larger gains with the increase in language model size. 
\vspace{-3.5mm}


\paragraph{Effects of prompts} To investigate the impact of prompt design on our method, we conduct experiments using a few hand-crafted prompts. We replace the words before the retrieved captions to probe the language model. The results are presented in Table~\ref{tab:prompt}. We find that there is no significant variation among the choices attempted, but some words such as {\tt Hints:}'' and {\tt Contexts:}'' perform slightly better on all datasets than the others, such as {\tt Subtitles:}'' and {\tt Captions:}''. Further optimization of the prompting words may potentially improve the performance.
\begin{table}[t]
    \centering
    \adjustbox{width=\linewidth}{
    \begin{tabular}{@{}ccccc@{}}
        \toprule
        Prompt & MSRVTT-QA & MSVD-QA & ANet-QA & TGIF-QA \\
        \midrule
        {\tt Subtitles:} & 17.3 & 36.6 & 26.2 & 51.1 \\
        {\tt Captions:} & 17.2 & 36.8 & 26.2 & 50.9 \\
        {\tt Hints:} & {\bf 18.0} & 36.8 & 26.3 & {\bf 51.2} \\
        {\tt Contexts:} & 17.6 & {\bf 36.9} & {\bf 26.4} & 51.1 \\
        \bottomrule
    \end{tabular}
    } 
    \vspace{-2mm}
    \caption{{\bf Effects of using different prompts.} We use WebVid-10M as the retrieval set and 20 retrieved sentences in all experiments.}
    \vspace{-6mm}
    \label{tab:prompt}
\end{table}
\vspace{-3.5mm}
\paragraph{Learning Visual-Text Projection for \shortmodelname} 
In Table~\ref{tab:ft_dataset_size}, we analyze the impact of multi-modal training on R2A. For efficiency purposes, we train R2A on various subsets of WebVid~\cite{frozenintime} for one epoch. Our findings reveal that while fine-tuning with relatively larger cross-modal data (\eg, 500K), there are some improvements on MSRVTT and TGIF. However, fine-tuning with smaller cross-modal data (\eg, 10k or 50k) can hamper the model's performance in the original setup. 
Future improvements may require dedicating more effort to architecture design and utilizing more training data. While beyond the scope of this study, we find this direction intriguing and encourage further research in this area.

\begin{table}[t]
    \centering
    \adjustbox{width=\linewidth}{
    \begin{tabular}{@{}lcccc@{}}
        \toprule
        Size & MSRVTT-QA & MSVD-QA & ANet-QA & TGIF-QA \\
        \midrule
        0 & 18.3 & {\bf 37.0} & {\bf 26.3} & 52.2 \\
        \midrule
        10k & 12.8 & 26.8 & 22.8 & 32.7 \\
        50k & 17.5 & 34.3 & 25.6 & 42.8 \\
        200k & 16.7 & 33.7 & 26.0 & 43.1 \\
        500k & {\bf 19.7} & 36.8 & 25.8 & {\bf 52.5} \\
        \bottomrule
    \end{tabular}
    } 
    \vspace{-2mm}
    \caption{{\bf Learning Visual-Text Projection for R2A.} In each experiment, we downsample the WebVid-10M dataset to the given size, and fix the number of training {\it epochs} to one.}
    \vspace{-5mm}
    \label{tab:ft_dataset_size}
\end{table}


\vspace{-2mm}
\section{Conclusions}
We propose \modelname, a framework for zero-shot VideoQA without task-specific training, utilizing off-the-shelf pre-trained multi-modal contrastive model. It transfers the zero-shot ability of a pre-trained LLM to a multi-modal setting without the need for explicitly learning video-language alignment. Specifically, we summarize the video modality with text via fast cross-modal retrieval in an external text corpus. Then, we probe a pre-trained language model with both the retrieved text and the question to predict the answer. Our design is highly flexible allowing easy component updates with no extra training. 
Experiments show that our R2A can achieve new state-of-the-art performance on multiple benchmarks.
\vspace{-2mm}
\section{Limitations}
The proposed Retrieving-to-Answer (R2A) approach is a promising direction for achieving zero-shot video question answering (VideoQA). Our attempt constitutes an important proof of concept in exploiting multimodal retrieval to enhance the generalization and robustness of current VideoQA frameworks. However, a major limitation of R2A is the extent to which the quality of the retrieved captions depends heavily on the performance of the retrieval model, as well as the diversity of the text corpus. Despite the remarkable abilities of CLIP in open-domain zero-shot cross-modal retrieval, it may still struggle to handle certain types of videos or text. Ideally, a sufficiently large dataset should encompass all topics and content of interest. In reality, however, there are still many cases where we cannot find the desired answers among existing data. Nevertheless, we posit that R2A constitutes a promising starting point and a baseline for future research on retrieval-based VideoQA.

\clearpage
{\small
\bibliographystyle{ieee_fullname}
\bibliography{egbib}
}

\end{document}